\documentclass{article}

\usepackage[final]{corl_2020} 
\usepackage{outlines}
\usepackage{graphicx}
\usepackage{amsmath}
\usepackage{amsfonts}
\usepackage{bbm}
\usepackage{authblk}

\usepackage{xcolor}
\usepackage{subcaption}
\usepackage{appendix}
\usepackage{algorithm}
\usepackage[noend]{algpseudocode}
\usepackage{relsize}
\usepackage{subcaption}
\usepackage{cleveref}
\usepackage{appendix}

\newcommand{\chrisnote}[1]{{\color{red} [[ #1 -- Chris ]] }}

\newcommand{\vladnote}[1]{{\color{blue} [[ #1 -- Vlad ]] }}

\newcommand{\wnote}[1]{{\color{purple} [[ #1 -- Willie ]] }}

\newcommand{\removed}[1]{}

\title{Amodal 3D Reconstruction for Robotic Manipulation via Stability and Connectivity}

\author[ ]{William Agnew}
\author[ ]{Christopher Xie}
\author[ ]{Aaron Walsman}
\author[ ]{Octavian Murad}
\author[ ]{Caelen Wang}
\author[ ]{Pedro Domingos}
\author[ ]{Siddhartha Srinivasa}
\affil[ ]{University of Washington}
\affil[ ]{\{wagnew3, chrisxie, awalsman, ovmurad, wangc21, pedrod, siddh\}@cs.washington.edu}

\begin{document}
\maketitle


\begin{abstract}
    Learning-based 3D object reconstruction enables single- or few-shot estimation of 3D object models. For robotics, this holds the potential to allow model-based methods to rapidly adapt to novel objects and scenes. Existing 3D reconstruction techniques optimize for visual reconstruction fidelity, typically measured by chamfer distance or voxel IOU. We find that when applied to realistic, cluttered robotics environments, these systems produce reconstructions with low physical realism, resulting in poor task performance when used for model-based control. We propose ARM, an amodal 3D reconstruction system that introduces (1) a stability prior over object shapes, (2) a connectivity prior, and (3) a multi-channel input representation that allows for reasoning over relationships between groups of objects. By using these priors over the physical properties of objects, our system improves reconstruction quality not just by standard visual metrics, but also performance of model-based control on a variety of robotics manipulation tasks in challenging, cluttered environments. Code is available at github.com/wagnew3/ARM.
\end{abstract}


\section{Introduction}

Manipulating previously unseen objects is a critical functionality for robots to ubiquitously function in unstructured environments. One solution to this problem is to use methods that do not rely on explicit 3D object models, such as model-free reinforcement learning \cite{schulman2017proximal, levine2018learning}. However, quickly generalizing learned policies across wide ranges of tasks and objects remains an open problem. On the other hand, obtaining detailed 3D object models can enable robots to physically reason about interactions with them to accomplish robotic tasks. For example, CAD models \cite{calli2015ycb} have extensively been used to detect the 6D pose of objects \cite{xiang2017posecnn, li2018deepim, tremblay2018deep}, facilitating many different kinds of manipulation tasks. Such 3D models can also be integrated with high-fidelity physics simulators \cite{coumans2016pybullet, todorov2012mujoco} to provide accurate simulations for planning and learning, enabling model-based methods\removed{ to work in realistic scenes and allowing robots} to generate high-level and/or low-level plans in order to accomplish long-horizon tasks \cite{dogar2011framework, stilman2007manipulation, williams2015model}. Unfortunately, these techniques can not be extended to unseen objects without building new models on the fly.

Generalizing interactive robotics problems to previously unseen objects using robust 3D reconstruction is the primary focus of this paper.  Rather than rely on a large database of models that have been laboriously hand-crafted or captured using a 3D scanner, we instead focus on techniques that can reconstruct meshes using observations of unseen objects in the robot's environment.  While SLAM methods can reconstruct highly accurate models given many views \cite{newcombe2011kinectfusion}, it can be challenging for these methods to separate objects in clutter and generate faithful reconstructions from a small number of observations.  The computer vision community has recently made significant progress in addressing these limitations using neural networks to estimate 3D models from single or few images \cite{gkioxari2019mesh, genre, smith2018multi, mescheder2019occupancy}. In this work, we investigate the use of such methods to solve robotic manipulation tasks involving previously unseen objects (instances and classes). 

\begin{figure}[t]
    \centering
    \includegraphics[width=\linewidth]{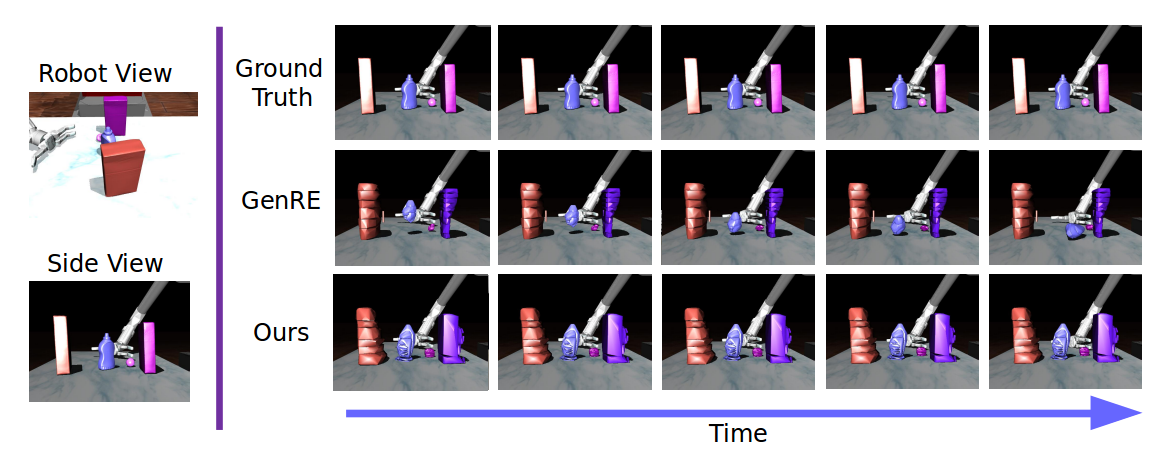}
    \caption{Comparison of physical behaviors of reconstructions from different algorithms. The baseline reconstruction of the light purple occluded mustard bottle is unstable and topples over, while our reconstruction is stable.} 
    \label{fig:teaser}
\end{figure}

Unfortunately, we find that directly applying state-of-the-art unseen object reconstruction techniques \cite{genre} to cluttered environments frequently fails to reconstruct objects in regions occluded by distractor objects, leading to physically unstable models and poor performance in downstream manipulation problems. This is due to these methods optimizing reconstruction metrics such as Chamfer distance, which are not necessarily relevant for manipulation tasks that utilize the resulting 3D model. Thus, our key insight is to adapt such systems to produce high \textit{physical} fidelity, improving manipulation success rates for unseen objects.

We accomplish this by encouraging the reconstruction network to provide physically realistic outputs. First, we assume that the scene and its objects are stable prior to manipulation, which motivates us to design a novel loss function that penalizes unstable reconstructions. This encourages the network to reconstruct stable scenes (see Figure~\ref{fig:teaser} for an example). Second, as mentioned above, current reconstruction methods struggle to adequately predict occluded portions of objects. This leads to disconnected objects which are not physically realistic. Thus, we design another loss function to penalize disconnectedness of predicted 3D models. Furthermore, both of our novel loss functions are differentiable which allows for end-to-end training. To our knowledge, we are the first to add physical priors on 3D reconstruction. Finally, we introduce a multi-channel voxel representation that allows reasoning over the spatial extent of other objects during the reconstruction phase, and we empirically show that this benefits performance. 

\removed{
We propose a solution to provide physically realistic reconstructions to allow the robot to accomplish the downstream manipulation task:
\begin{enumerate}
    \item We assume that objects and scenes are stable prior to manipulation. We introduce a novel differentiable loss function that penalizes unstable reconstructions, thus encouraging stable scenes. See Figure~\ref{fig:teaser} for an example.
    \item Objects are connected. We introduce a novel differentiable loss function that penalizes disconnectedness to reconstruct connected objects. \chrisnote{Willie, can you help motivate this? Not sure what your thoughts are on this. Maybe this helps more in occluded regions?}
    \item Reconstruction requires reasoning not only about the object being reconstructed, but also about how it interacts with other objects in the scene. Thus, we introduce a multi-channel scene representation that allows reasoning over relationships between objects when reconstructing.
\end{enumerate}
}

We integrate our proposed loss functions into a modular framework to provide Amodal 3D Reconstructions for Robotic Manipulation (ARM). We use the state-of-the-art method, GenRE~\cite{genre}, as our reconstruction method, however we are free to choose any method in place of GenRE as our framework is modular. To evaluate our method, we introduce a challenging clutttered 3D reconstruction benchmark. We empirically demonstrate that ARM improves the reconstruction quality by 28\% on this task, and manipulation success rates on unseen objects on a range of challenging tasks including grasping, pushing, and rearrangement by 42\% over GenRE.

\section{Related Work}

\textbf{3D Reconstruction.}
3D reconstruction is a challenging problem that has been studied for decades. Recently, learning-based methods have provided significant progress when focusing on reconstructing single objects in isolation \cite{tulsiani2018factoring, nie2020total3dunderstanding}. Recently, \citet{gkioxari2019mesh} introduces graph neural networks to refine mesh predictions. \citet{kulkarni20193d} introduces pairwise object relations and a refinement procedure to improve object pose and shape estimation. Additionally, reconstructing previously unseen classes compounds the difficulty of the problem \cite{yan2016perspective, shin2018pixels, genre}.

Amodal 3D reconstruction is the problem of reconstructing partially visible objects, which is still a relatively unexplored task \cite{han2019image}. \citet{kundu20183d} approaches this problem by learning class-specific shape priors from large datasets of CAD models. \citet{kulkarni20193d} study amodal reconstruction of scenes from single RGB images, while \citet{sucar2020neural} handles occlusion by using multiple RGBD views.
However, because robot manipulation settings are our desired environment, we require not only amodal reconstruction of objects, but also the ability to deliver physically realistic reconstructions which warrants more informed loss functions including stability and connectivity. 

\textbf{Exploiting Physics for Scene Understanding}
Some works have investigated the use of physics to better inform reconstructions by encouraging physical plausibility. In particular, \cite{jia20133d, du2018learning} use a stability prior and \citet{chen2019holistic++} use collision and support priors to fit 3D bounding boxes around objects. Our work introduces a differentiable and efficiently computable stability prior to allow generation of stable 3D meshes, rather than just 3D bounding boxes. Additionally, our connectivity prior promotes better reconstruction in occluded regions.

\textbf{3D Reconstruction in Robotics}
While applying 3D reconstruction to robotics provides an appealing solution to manipulation, few works have investigated this. Such reconstructions can be used to synthesize grasps for single objects using analytic and/or learning-based solutions \cite{varley2017shape, van2019learning, yan2018learning}. \citet{sucar2020neural} considers grasp synthesis in tabletop scenes with multiple objects of known classes, but does not consider highly cluttered scenes. Most similar to our work, \cite{lundell2019robust, lundell2019beyond} compute grasps for reconstructed objects in cluttered scenes. However, they do not take advantage of physics, which reduces the physical realism. Our work attempts to solve a wider range of manipulation tasks while incorporating physical notions of stability and connectivity to improve performance.


\section{Amodal 3D Reconstruction}
\begin{figure}
\centering
\includegraphics[width=1\textwidth]{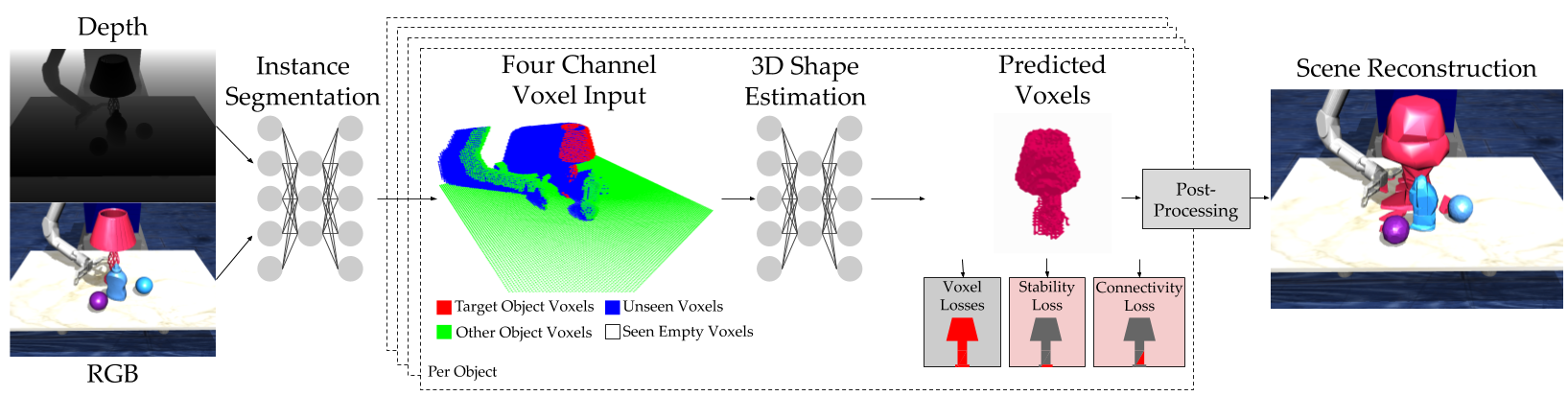}
\caption{A visual representation of our modular framework.}
\label{fig:stability}
\end{figure}

\subsection{System Architecture}

In this section we describe the architecture of our ARM framework, which consists of four stages. 1) We first apply an instance segmentation network to the input RGB-D image. 2) For each object we detect, we pre-process its point cloud to compute its \textit{four channel voxel input representation}, defined below. 3) ARM uses this representation to perform 3D shape estimation with a deep network, followed by post-processing. 4) Lastly, we obtain mesh representations which we employ for manipulation planning. Our framework is visually summarized in Figure~\ref{fig:stability}.

\textbf{Instance Segmentation} ARM takes as input a RGB image, $I \in \mathbb{R}^{h \times w \times 3}$, and an organized point cloud, $P \in \mathbb{R}^{h \times w \times 3}$ computed by backprojecting a depth image with camera intrinsics. 
This is passed to an instance segmentation network $\mathcal{S}$ which outputs instance masks $L = \mathcal{S}(I, D)\in \mathcal{L}^{h \times w}$, where $\mathcal{L} = \{0, \dots, K\}$ and $K$ is the number of detected object instances.
We use UOIS-Net \cite{xie2019best} as $\mathcal{S}$ which produces high quality segmentations for unseen objects.

\textbf{Four Channel Voxel Input Computation} We introduce a four-channel voxel representation to enable ARM to reason about the spatial extent of other objects during reconstruction. For each object $o \in \mathcal{L}$, we compute a voxel occupancy grid $F_o \in \{0, 1\}^{d^3 \times 4}$ augmented with the surrounding objects' occupancies, as well as with voxel visibilities with respect to the camera. 
Let $F_{o}^i$ denote the $i^{\textrm{th}}$ channel of $F_o$.
$F_o^{1}$ is the voxel grid of object $o$ alone, which is computed by voxelizing $P_o$, the point cloud segmented with the instance mask for $o$. $F_o^2$ contains all other objects in $L$ except for $o$. $F_o^3$ consists of a mask of empty voxels, and $F_o^4$ contains unobserved voxels. Note that $F_o^3, F_o^4$ are computed using the camera extrinsics and intrinsics. $F_o$ is centered at the center of mass of object $o$ and has side length $k\delta_o$, where $\delta_o$ is the maximum distance between points in $P_o$. In our implementation, $k = 4$ to allow filling of occluded regions.
Finally, we translate $F_o$ so the table occupies the $z=0$ plane in our voxel grid.
\removed{ by finding the $z$ plane of height $z_{table}$ in $P \setminus P_o$ with the most set voxels and shift $F_{o}$ downwards by $z_{table}$\vladnote{Why is the table plane computed for each object. If it is not, it seems that way from this paragraph} \wnote{it is computed per object--objects could be on different (or multi-level) tables}}

\textbf{3D Shape Estimation and Scene Reconstruction} For each object $o \in \mathcal{L}$, we use $F_{o}$ as input to a 3D reconstruction network $\mathcal{C}$ which outputs the probability of $o$'s presence at each voxel as $\mathcal{C}(F_o)=V_o \in [0,1]^{d^3}$. We use GenRe-Oracle \cite{genre} as our 3D reconstruction network\removed{ increasing the channels on the input convolutions to handle the additional channels in our input representation}. GenRe-Oracle is a modfication of GenRe that uses depth, rather than RGB data. GenRe-Oracle projects observed pointclouds onto a sphere, inpaints them using a CNN, backprojects to a 3D voxel grid, and then refines the voxel grid using a 3D encoder-decoder architecture. Finally, we use marching cubes \cite{lorensen1987marching} after thresholding $V_o$ to transform the output voxel probabilities into a mesh. 

We post-process the meshes in order to make them suitable for physics simulation. First, we remove intersections between meshes to prevent inconsistent behavior in simulation, removing from the larger of the intersecting meshes. We then compute an approximate convex decomposition of each mesh using V-HACD \cite{mamou2016volumetric}.

\textbf{Manipulation Planning}
We pose the task of manipulation planning in the form of an MDP consisting of an action space $a \in A$, a state space $s \in S$, a stochastic transition function $G(s', s, a)=P(s'|s, a)$, and a single RGB-D view of the corresponding environment. We solve this MDP by reconstructing every object in the view, instantiating a simulation of the environment from the robot viewpoint with the reconstructed objects, and then using a physics simulator \cite{todorov2012mujoco} to approximate the transition function $G$. We use MPPI \cite{williams2015model} to plan a sequence of actions in the simulator, and execute this plan in the real environment.


\subsection{Loss Functions}
\begin{figure}[t]
\centering
\includegraphics[width=1\textwidth]{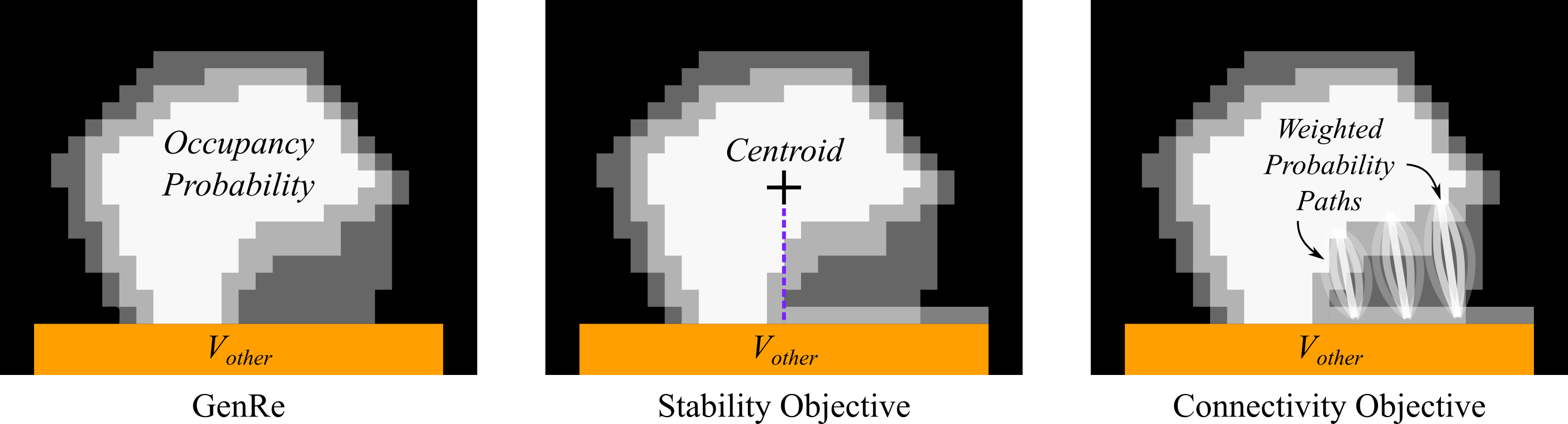}
\caption{Impact of stability and connectivity objectives. Left: occupancy probabilities of an estimated shape, in greyscale. Adding the stability objective makes the object stable, and adding the connectivity objective fills in the gap between the shape and inferred base.}
\label{fig:objectives}
\end{figure}

GenRE~\cite{genre} uses a weighted combination of cross entropy and a surface loss between reconstructed and ground truth voxels during training. However, in robotic settings, optimizing these losses alone are not sufficient to solve the downstream task of robotic manipulation, as we show in Section \ref{robot_exps}. This results in reconstructions with poor physical fidelity during the planning phase, often due to instability of poor reconstruction of occluded regions. We tackle this issue by designing auxiliary differentiable loss functions based on two physical priors: 1) objects are stable prior to manipulation, and 2) objects are a single connected component. Figure \ref{fig:objectives} gives an overview of these loss functions.


\subsubsection{Stability Loss}
Our stability loss provides a prior over object shape, even in occluded regions, by reasoning about hidden supports objects may have. An object is in static equilibrium if the net forces acting upon it are equal to zero \cite{urone_dirks_sharma}. This means that the center of mass is within the base of support of an object. Technically, the center of mass must be behind a pivot point (where the object rests on another object) along every direction $s$ perpendicular to the force of gravity $\vec{g}$.

We first define some notation here.
Recall that $V_o$ parameterizes a multivariate Bernoulli distribution over binarized voxel grids. 
For sample $v \sim V_o$, we define $M(v)$ to be the center of mass of $v$.
Furthermore, let $i \in d^3$ index the voxel grid, and $S = \{s : s \perp \vec{g}\}$ be the set of directions perpendicular to $\vec{g}$. Then, for each $s \in S$, let $i^s$ and $M^s(v)$ be the projections of $i$ and $M(v)$ onto the plane defined by $s$ and $\vec{g}$ that passes through the origin. 
We denote $H_s(i)$ as the set of voxels belonging to other objects that support $i$ in direction $s$, which can happen when $i$ is directly above or leaning against such voxels. Finally, $V_{\bar{o}}$ is the probabilities of other objects output by the 3D reconstruction network. 

Given this notation, we can define our stability loss to be the probability that $v$ is stable. Let $E(v)$ be the event that $v$ is stable. Then our stability objective is defined as

\begin{gather}
\label{eq:stability_loss}
P(E(v))=\prod_{s \in S} (1-u_s)\\
u_s=\prod_{i \in d^3} \Big[ 1- V_o(i) P\Big(i^s>M^s(v)\Big) h_s(i) \Big], \ \ \ h_s(i)=1-\prod_{i' \in H_s(i)} \Big( 1-V_{\bar{o}}(i') \Big)
\end{gather}

$u_s$ is the probability that $v$ is unstable in direction $s$. It is the probability that every voxel $i$ is unstable; that is $i$ either doesn't exist, doesn't support $v$ along direction $s$, or isn't supported ($h_s(i)$). Eq. (\ref{eq:stability_loss}) is intractable, so in order to take the gradient we introduce independence assumptions and the approximation that a voxel $i$ exists only if $V(i) \geq 0.5$ to derive an efficiently computable derivative of object stability with respect to each object voxel:

\begin{gather}
    \frac{d \log P(E(v))}{d V_o(i)}=\sum_{s \in S}\frac{-u'_s}{1-u_s} \\
    u'_s=-P\Big(i^s>M^s(v)\Big)  \hat{h}_s(i) \prod_{i_o \in d^3, i_o \neq i} \Big[ 1-P\Big(i_o^s>M^s(v)\Big) \mathbbm{1}\{V(i_o) \geq 0.5\} \hat{h}_s(i_o) \Big] \\
    \hat{h}_s(i_o)=1-\prod_{i_b \in H_s(i)} \Big[ 1- \mathbbm{1}\{V_{\bar{o}}(i_b) \geq 0.5\} \Big]
\end{gather}





This gradient captures several intuitive properties of stability. If an object has even a single voxel supporting it in a particular direction then it is stable. For a direction $s$, if a single supported voxel $i_o$ exists, then $u'_s$ is close to zero, and the magnitude of the derivative in that direction will be small. Vice versa, when no supporting voxel is present, $u'_s$ is close to 1 and the magnitude is nontrivial. 
This captures the idea that when supporting voxels are present, the effect on stability is small, but when no supporting voxels are present, the effect is large. 
Importantly, this prior is shape agnostic: it is not biased towards making an object stable by adding a base under existing voxels, for example, but rather only increases the probability of any voxel that would make the object stable, minimizing reconstruction deviation from the learned shape prior. A full derivation can be found in appendix A.

\subsubsection{Connectivity Loss}
Our connectivity loss imposes a prior on object shape even in occluded regions by allowing the network to infer connections between disjoint parts of observed objects. This complements the stability objective which frequently infers occluded bases of objects.\removed{ Let voxels be 26-connected.} We define $v$ to be connected if for every pair of existent voxels $a, b$, there exists a path $t=\{i_0, i_1, \dots \}$ between $a$ and $b$. The probability that a path $t$ exists in $v$ is $P(t)=\prod_{i \in t} V_o(i)$. Let $T(a, b)$ be the set of all possible paths between $a$ and $b$, $C(v)$ be the event that $v$ is connected, and $C(a, b)$ be the event that there is a path between $a$ and $b$. Then we define our connectivity objective as

\begin{equation}
\label{eq:connectivity_loss}
P(C(v))=\prod_{a, b \in d^3, a\neq b} \Big[ V_o(a)V_o(b)P(C(a,b))+1-V_o(a)V_o(b) \Big]
\end{equation}


The derivative of this equation is intractable because it requires considering every path $t$ between every vertex pair $(a, b)$. To resolve this, we note that relative to the most likely path $t^*$ between $a$ and $b$, most paths have small probability. Thus, for any other voxel $c$, we may ignore low probability paths passing through $c$ when calculating their contribution to the connectivity of $a$ and $b$ and only consider the most likely path from $a$ to $b$ passing through $c$. With this approximation, our per-voxel derivative of Eq. (\ref{eq:connectivity_loss}) is

\begin{equation}
\frac{d\log P(C(v))}{d V_o(c)}=\sum_{a, b \in d^3, a \neq b \neq c} \frac{V_o(a) V_o(b) \frac{d}{d V_o(c)}P(C(a, b))}{V_o(a) V_o(b) P(C(a, b))+1-V_o(a) V_o(b)}
\end{equation}


\removed{
\[
    \frac{d P(\texttt{connected}(a, b))}{dV(i)}= 
\begin{cases}
    \frac{P(t^s)}{V(i)}, & t^s=t^i\\
    \frac{P(t^i)}{V(i)}(1-P(t^s)), & t^s \neq t^i\\
\end{cases}
\]
}
where $P(C(a, b)) = P\left(\bigcup_{t \in T(a,b)}\ t\right) \approx P(t^* \cup t^c)$, $t^*$ is the path from $a$ to $b$ with the highest probability of existing, and $t^c$ is the path from $a$ to $b$ that includes $c$ with the highest probability of existing. This approximation preserves several desirable properties of the exact gradient. First, it only encourages connecting the object by reinforcing the most likely paths, rather than the physically shortest paths. By considering each most likely path from $a$ to $b$ that passes through $c$, it also produces dense connections, rather than only amplifying the shortest path between $a$ to $b$, which would often result in shapes connected by single voxel width paths. 

\section{Experiments}

\begin{figure}[t]
\centering

\begin{subfigure}[t]{.5\textwidth}
\centering
\includegraphics[width=\textwidth]{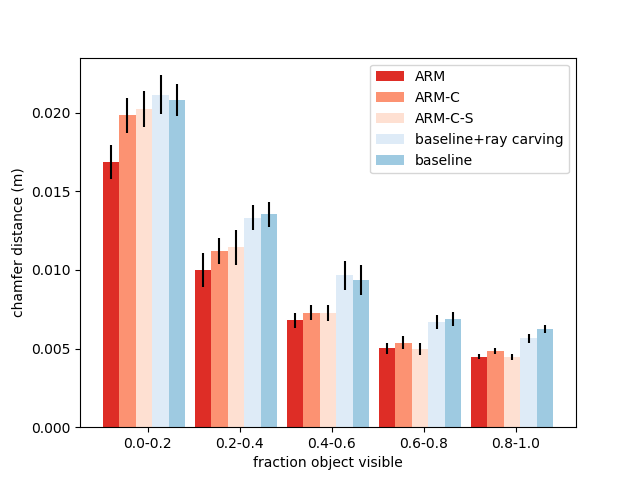}
\captionsetup{width=0.9\textwidth}
\end{subfigure}%
\hfill
\begin{subfigure}[t]{.5\textwidth}
\centering
\includegraphics[width=\textwidth]{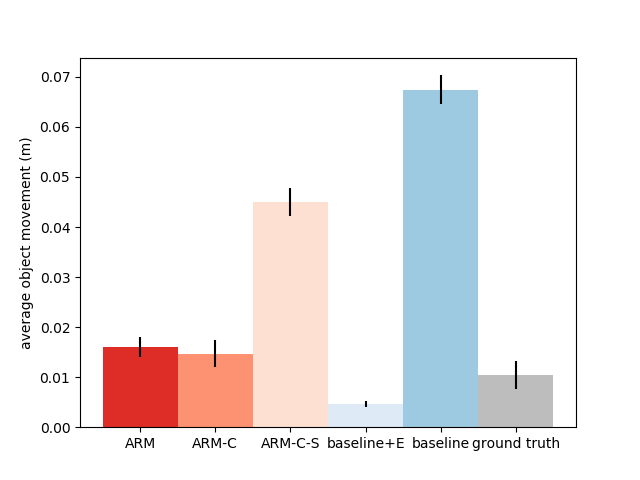}
\captionsetup{width=0.9\textwidth}
\end{subfigure}
\caption{(left) Chamfer distances on held-out objects, broken down by observation occlusion. (right) Average stability of reconstructed objects. Error bars are a 90\% confidence interval.}
\label{fig:reconstruction_quality}
\end{figure}

\subsection{Implementation Details}
We implement ARM using UOIS-Net \cite{xie2019best} for instance segmentation, and the GenRE depth backbone \cite{genre} for 3D reconstruction. We use MuJoCo \cite{todorov2012mujoco} as a physics simulator for our reconstructed environment. 
To train ARM, we create a large dataset of cluttered tabletop scenes in MuJoCo using ShapeNet \cite{chang2015shapenet} tables and objects. We divide the ShapeNet objects into training and test sets, containing 4803 and 3368 unique objects respectively. For each scene, we drop between 5 and 20 randomly selected objects onto a table to ensure cluttered scenes and stacked objects with complex stability relationships. We render several views with randomized camera positions, using a custom OpenGL renderer to produce realistic images. Each network is trained with ADAM for approximately 100,000 iterations with a batch size of 16. Stability and connectivity loss gradients are only applied on occluded voxels, as all other voxels are observed to be either empty or occupied. Additional implementation and training details are in Appendix C.

\subsection{Baselines}

\begin{figure}
\centering
\includegraphics[width=1\textwidth]{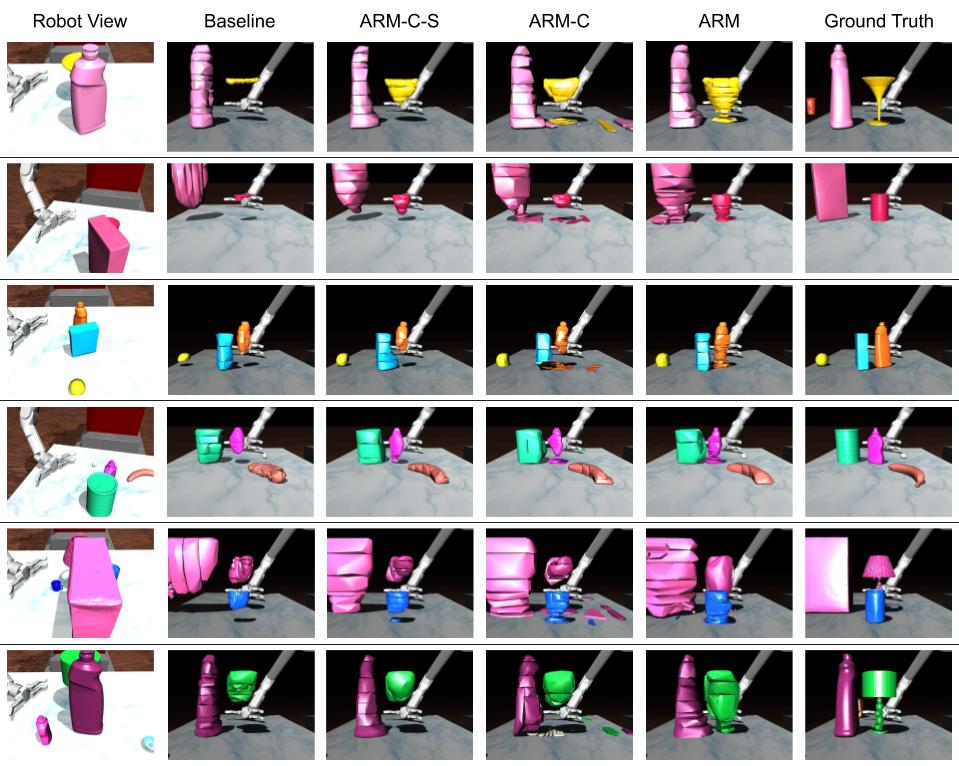}
\captionsetup{width=1\textwidth}
\captionof{figure}{Qualitative reconstruction results. ARM is able to infer both bases and occluded object regions.}
\label{fig:qualitative_recon}
\end{figure}

Our main baseline that we compare against is GenRE-Oracle \cite{genre}, which we denote as baseline. In order to test the most direct way of using the information about observed occupied and unoccupied voxels encoded in the four channel representation, we introduce a simple modification to GenRE to give baseline+ray carving where we remove all observed empty voxels after reconstruction. Additionally, in order to test their significance, we train two ablations of our method, ARM-C, ARM without the connectivity prior and ARM-C-S, ARM without the connectivity or stability prior.


\subsection{Reconstruction Quality}
We quantitatively compare the visual reconstruction quality of ARM to our baselines on reconstruction of cluttered scenes generated with held-out test objects and ground truth segmentations in Figure \ref{fig:reconstruction_quality} (left). ARM outperforms the baseline at all occlusion levels, improving Chamfer loss by 28\% overall. This improvement is especially pronounced on highly occluded objects, where the stability and connectivity objectives combined allow ARM to estimate occluded bases and fill gaps between those bases and the observed parts of objects. However, low Chamfer distance alone is not sufficient for accurate simulation; reconstructed objects must also exhibit similar physics to the ground truth. In Figure \ref{fig:reconstruction_quality} (right), we measure scene stability by placing each reconstructed scene into a physics engine, simulating forward for five seconds with gravity as the only force, and measuring the $L_2$ displacement of the reconstructed object centers. As a simple yet effective baseline for stability, we consider baseline+E, where we extrude the reconstructed mesh (from baseline) down to the table to ensure mesh stability. Both the baseline and ARM-C-S frequently reconstruct unstable objects that fall or tumble. Adding the stability prior improves object stability to near that of the ground truth meshes. Note that ground truth meshes move a small amount because Mujoco considers only one point of contact between each pair of mesh geometries, which can cause meshes to slowly move.

In Figure \ref{fig:qualitative_recon}, we provide qualitative reconstruction results. The baseline and ARM-C-S only reconstruct the visible portions of occluded meshes at the top, producing reconstructions that will tumble over as soon as simulation begins. ARM-C reconstructs the bases to produce stable meshes, but still leaves large voids in the middle of reconstructions which frequently cause manipulation to fail. On the other hand, ARM is able to both reconstruct bases and effectively reason about occluded regions, producing a tapered reconstruction for the yellow wine glass in the top row, but filling in a cylinder for the blue tin can in the fifth row.

\subsection{Robot Manipulation} \label{robot_exps}
\textbf{Robot Manipulation Tasks}
To evaluate the efficacy of our method on robot manipulation tasks, we create a suite of robotics manipulation tasks across a range of challenging objects in cluttered scenes. We consider three important robot tasks: grasping, pushing, and rearrangement, which entails grasping and pushing/pulling. In each task, the robot first creates a 3D reconstruction of the environment from an RGBD observation. It then plans a trajectory with the 3D reconstruction, and finally executes the trajectory in the ground truth environment. We execute each task on 14 different target objects from the YCB dataset \cite{calli2015ycb} and 12 from a set of challenging, highly non-convex objects downloaded from online 3D repositories, all previously unseen during training (see Figure \ref{fig:qualitative_recon} for examples). For each task and target object we consider 10 occlusion intervals, or fraction of the target manipulation object visible to the robot, from $[0,0.1)$ to $[0.9,1)$, and no occlusion. For each task, target object, and occlusion interval, we generate three scenes by randomly placing unseen YCB objects until the occlusion of target manipulation object is within the desired range, for total of 2574 tasks. To isolate the effects of different 3D reconstruction algorithms, we use ground truth instance segmentations. More details on our cluttered robot manipulation benchmark are available in Appendix D.

\textbf{Robot Task Performance}
Figure \ref{fig:manipulation_results} (left) shows average task success rates on the manipulation tasks for each of the methods. ARM achieves the best performance across all tasks, improving the baseline and the extrusion baseline by 42\% and 25\% respectively. In Figure \ref{fig:manipulation_results} (middle), we break down performance by target object occlusion, and find that while ARM and the baseline perform similarly at very low levels of occlusion, at almost all levels of visibility below 80\% ARM performs the best. Notably, while the baseline success rate on 10\% visible objects is only about 25\% the success rate on completely unoccluded objects, ARM's success rate on 10\% visible objects is 75\% its success rate on unoccluded objects, showing that our stability and connectivity priors are less prone to performance degradation in the face of occlusion, which is due to our stability and connectivity priors. While ARM gives significant improvements over the baseline, this task suite is still quite challenging: most tasks involve high levels of target object occlusion, and even poor reconstructions of self-occluded object regions can cause manipulation failure.
Lastly, we analyze the impact of object stability on performance in Figure \ref{fig:manipulation_results} (right), where we plot task success vs. target object stability, showing a correlation between task success and object stability. 

\begin{figure}
\centering

\begin{subfigure}[t]{.33\textwidth}
\centering
\includegraphics[width=1\textwidth]{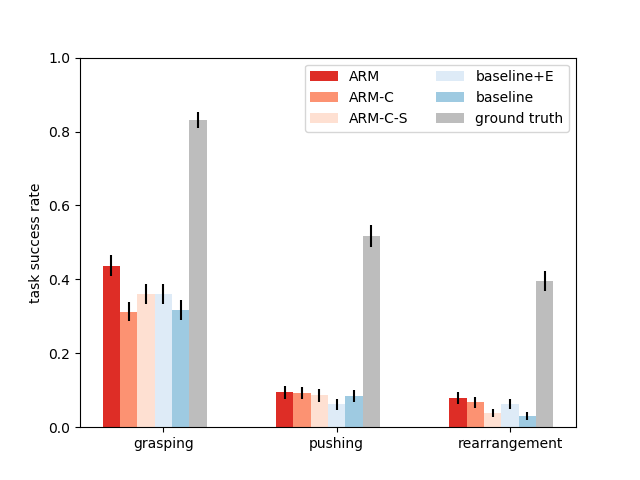}
\end{subfigure}%
\hfill
\begin{subfigure}[t]{.33\textwidth}
\centering
\includegraphics[width=1\textwidth]{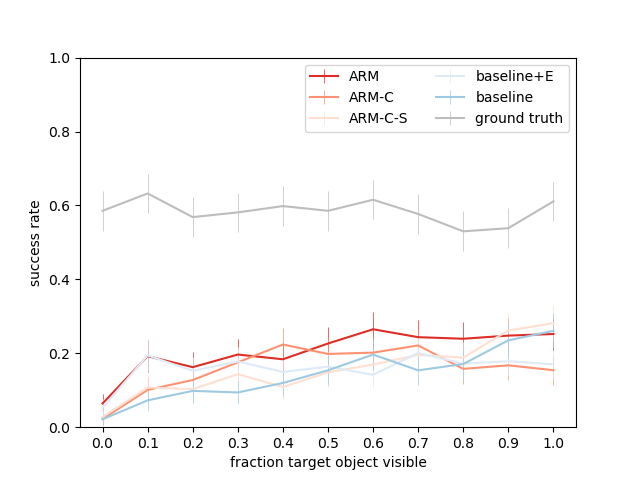}
\end{subfigure}%
\hfill
\begin{subfigure}[t]{.33\textwidth}
\centering
\includegraphics[width=1\textwidth]{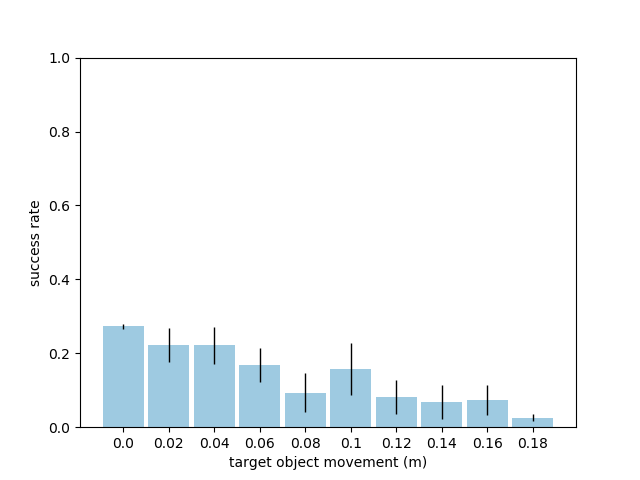}
\end{subfigure}%

\caption{(left) Success rates on manipulation tasks using models generated by different reconstruction algorithms. (middle) Manipulation success rate vs. target object occlusion. Visibilities are binned in increments of 0.1, so 0.0 includes all visibilities in [0,0.1). (right) Task success vs. target object stability. Error bars are a 90\% confidence interval.}
\label{fig:manipulation_results}
\end{figure}

\section{Conclusion}
We have shown that directly applying 3D object reconstruction methods in cluttered robotic environments can produce reconstructions with low physical fidelity, which often leads to unsuccessful task execution. We proposed a modular framework, ARM, that includes a stability prior, a connectivity prior, and a multi-channel input representation to deliver more physically faithful reconstructions. ARM generates 3D reconstructions that are not only better by standard visual loss metrics, but more importantly they allow for significantly better robot task performance in challenging cluttered scenes. 
We hope our reconstruction system will enable further model-based learning and control applications. While ARM enables significant improvements over the baseline on reconstruction and manipulation of occluded objects, performance on highly occluded objects is still far from that of the ground truth. To enable further research on this challenging task, we will publicly release our large, high-quality cluttered dataset and robot evaluation benchmarks.

\section{Acknowledgements}
We would like to thank the authors of UOIS~\cite{xie2019best}, GenRE~\cite{genre}, Dex-Net 2.0~\cite{mahler2017dex} and POLO~\cite{Lowrey-ICLR-19} for releasing code used in this paper. We would like to thank Jonathan Tremblay for help on early versions of this project. We would like to thank Aravind Rajeswaran, Kendall Lowrey, and Colin Summers for advice and help with MuJoCo. This work was supported by an NDSEG Fellowship and ONR grant N00014-18-1-2826. The GPU machine used forthis research was donated by Nvidia. This work was partially funded by the National Institute of Health R01 (\#R01EB019335), National Science Foundation CPS (\#1544797), National Science Foundation NRI (\#1637748), the Office of Naval Research, the RCTA, Amazon, and Honda Research Institute USA.

\bibliography{example}  

\begin{appendices}

\section{Stability Loss Derivation}
Recall that $V_o$ parameterizes a multivariate Bernoulli distribution over binarized voxel grids. 
For sample $v \sim V_o$, we define $M(v)$ to be the center of mass of $v$.
Furthermore, let $i \in d^3$ index the voxel grid, and $S = \{s : s \perp \vec{g}\}$ be the set of directions perpendicular to $\vec{g}$. Then, for each $s \in S$, let $i^s$ and $M^s(v)$ be the projections of $i$ and $M(v)$ onto the plane defined by $s$ and $\vec{g}$ that passes through the origin. 
We denote $H_s(i)$ as the set of voxels belonging to other objects that support $i$ in direction $s$, which can happen when $i$ is directly above or leaning against such voxels. Finally, $V_{\bar{o}}$ is the probabilities of other objects output by the 3D reconstruction network. 

Given this notation, we can define our stability loss to be the probability that $v$ is stable. Let $E(v)$ be the event that $v$ is stable. Then

\[P(E(v))=P(\texttt{$v$ stable along all directions}) \]

Here we introduce our first approximation by discritzing the set of all directions into a finite set of evenly space directions $S$. In our implementation, we used $|S|=25$. Then

\[P(E(v))=\prod_{s_i \in S} P(\texttt{$v$ stable along $s_i$| $v$ stable along $s_0, s_1, ..., s_{i-1}$})\]

We introduce a second approximation, that stability along different directions is independent. Then

\[P(E(v))=\prod_{s_i \in S} P(\texttt{$v$ stable along $s_i$})=\prod_{s_i \in S}1-P(\texttt{$v$ not stable along $s_i$})=\prod_{s_i \in S} 1-u_s \]
\[u_s=P(\texttt{no voxel $i \in d^3$ makes $v$ stable}=\prod_{i \in d^3} 1- P(\texttt{i makes $v$ stable}) \]
\[=\prod_{i \in d^3} 1- P(i^s>M^s(v))P(\texttt{$i$ is supported by another voxel in direction $s$}) \]
\[P(E(v))=\prod_{s_i \in S}\prod_{i \in d^3} \Big[ 1- V_o(i) P(i^s>M^s(v)) h_s(i) \Big], \ \ \ h_s(i)=1-\prod_{i' \in H_s(i)} \Big( 1-V_{\bar{o}}(i') \Big) \]

$P(i^s>M^s(v))$ is just the cdf of $M^s(v)$, which is a linear combination of Bernoulli variables. This is a generalization of the Poisson Binomial distribution, the cdf of which is inefficient to compute exactly [cite].
Therefore, we use a normal approximation of the cdf over the sum of weighted Bernoulli variables:

\[P(i^s>M^s(v)) \approx \Phi_{\mu, \sigma}(i^s),\ \ \ \ \mu=\sum_{i \in d^3} i^{s}v(i), \ \ \ \ \sigma=\Big[\sum_{i \in d^3} (i^s_{v})^2 v(i)(1-v(i)) \Big]^{1/2} \]

Now we compute the derivative of $P(E(v))$ with respect to voxel $i$. First, we take the log of $[P(E(v))$, noting that maximizing the log also maximizes $[P(E(v))$. Then

\[\frac{d \log P(E(v))}{d V_o(i)}=\sum_{s \in S}\frac{-u'_s}{1-u_s} \]
\[u'_s=-P\Big(i^s>M^s(v)\Big)  h'_s(i) \prod_{i_o \in d^3, i_o \neq i} \Big[ 1-P\Big(i_o^s>M^s(v)\Big) V(i_o) h'_s(i_o) \Big] \]
\[ \hat{h}_s(i_o)=1-\prod_{i_b \in H_s(i)} \Big[ 1- V_{\bar{o}}(i_b) \Big] \]

Note that 

\[\frac{d \log P\Big(i^s>M^s(v)\Big)}{d V_o(i)}=0\]

as adding (or increasing the chance that a number appears) in a set of numbers cannot change whether the mean of that set of numbers is greater than, equal to, or less than that number.

The derivative is numerically unstable since the product $\prod_{i_o \in d^3, i_o \neq i} \Big[ 1-P\Big(i_o^s>M^s(v)\Big) V(i_o) h'_s(i_o) \Big]$ has millions of terms in practice and will therefore often underflow to zero. To fix this we note that, during inference, if any voxel has $v(i) \geq 0.5$ then we set it as existing, and the notion of continuously existing voxels is only an approximation to aid differentiation. Therefore, we may write

\[ \frac{d \log P(E(v))}{d V_o(i)}=\sum_{s \in S}\frac{-u'_s}{1-u_s} \]
\[ u'_s=-P\Big(i^s>M^s(v)\Big)  \hat{h}_s(i) \prod_{i_o \in d^3, i_o \neq i} \Big[ 1-P\Big(i_o^s>M^s(v)\Big) \mathbbm{1}\{V(i_o) \geq 0.5\} \hat{h}_s(i_o) \Big] \]
\[ \hat{h}_s(i_o)=1-\prod_{i_b \in H_s(i)} \Big[ 1- \mathbbm{1}\{V_{\bar{o}}(i_b) \geq 0.5\} \Big] \]

completing our derivation of the stability loss.

\section{Connectivity Loss Derivation}

Our connectivity loss imposes a prior on object shape even in occluded regions by allowing the network to infer connections between disjoint parts of observed objects. This complements the stability objective which frequently infers occluded bases of objects.\removed{ Let voxels be 26-connected.} We define $v$ to be connected if for every pair of existent voxels $a, b$, there exists a path $t=\{i_0, i_1, \dots \}$ between $a$ and $b$. The probability that a path $t$ exists in $v$ is $P(t)=\prod_{i \in t} V_o(i)$. Let $T(a, b)$ be the set of all possible paths between $a$ and $b$, $C(v)$ be the event that $v$ is connected, and $C(a, b)$ be the event that there is a path between $a$ and $b$. Then we define our connectivity objective as the probability that, for every pair of vertices $a,b$, a exists and b exists and a and b are connected, or not (a exists and b exists), conditioned on the probabilities. We introduce the approximation that voxel connectivity is independent between pairs of voxels. Then

\[P(C(v))=\prod_{a, b \in d^3, a\neq b} \Big[ V_o(a)V_o(b)P(C(a,b))+1-V_o(a)V_o(b) \Big] \] Taking the log and then derivative with respect to a voxel $c$ we get

\[ \frac{d\log P(C(v))}{d V_o(c)}=\sum_{a, b \in d^3, a \neq b \neq c} \frac{V_o(a) V_o(b) \frac{d}{d V_o(c)}P(C(a, b))}{V_o(a) V_o(b) P(C(a, b))+1-V_o(a) V_o(b)} \]

Computing $P(C(a,b))$ requires us to compute combinatorially many paths and is infeasable. We note that many paths are highly unlikely and contribute little to this probability. Therefore, we introduce the approximation of computing $P(C(a,b))$ using the most likely $a,b$ path and the most likely $a,b$ path that includes $c$; that is

\[ P(C(a, b))=\underset{t \in T(a,b)} {\mathlarger{\mathlarger{\lor}}} P(t) \approx P(t^* \lor t^c)\]

\[P(t^* \lor t^c)= \begin{cases}
    P(t^*), & t^*=t^c\\
    1-(1-P(t^*))(1-P(t^c)), & t^* \neq t^c\\
\end{cases} \]

and 

\[
    \frac{d P(C(a, b))}{dV(i)}= 
\begin{cases}
    \frac{P(t^s)}{V(i)}, & t^s=t^i\\
    \frac{P(t^i)}{V(i)}(1-P(t^s)), & t^s \neq t^i\\
\end{cases}
\]

completing our derivation.

\section{System Implementation Details}

Our system is implemented in PyTorch. We train using ADAM for 104,000 iterations with a batch size of 16 and a learning rate of 2e-4. Other training parameters are identical to those used in GenRE. We coarsen the predicted voxel grid by a factor of 8 during connectivity loss computation for efficiency.

\section{Cluttered Reconstruction Benchmark Details}
We introduce a cluttered reconstruction benchmark containing 2318794 training and 484161 test reconstruction instances of cluttered tabletop objects. Objects are drawn from the shapenet xyz categories. Scene generation uses the following randomization parameters:

\begin{center}
\begin{tabular}{| c | c |}\hline
 table & sampled randomly from shapenet   \\ \hline
 table objects & sampled from selected shapenet categories [1]  \\ \hline
 number table objects & uniform(3,20) \\ \hline
 table and object color & uniform random \\ \hline
 object scale & uniform(0.5, 4) \\ \hline
 object rotation & \begin{tabular}{@{}c@{}}75\% chance of being none/default (upright object) \\ 25\% chance of being uniform random\end{tabular} \\ \hline
 object drop position center & uniform random point on table \\ \hline
 object drop position std dev & 0.1m \\ \hline
 object drop height & table height+0.2m \\ \hline
 camera position x,y & uniform in shell between 0.125 and 1.5m centered on table \\ \hline
 camera position z & \begin{tabular}{@{}c@{}}uniform in half-shell between 0.125 and 1.5m \\ centered on table, on top of table\end{tabular} \\ \hline
 camera lookat & \begin{tabular}{@{}c@{}}random point in sphere of radius \\ min(camera distance from table/2, 0.25)\end{tabular} \\ \hline
 \end{tabular}
\end{center}

Parameters are selected according to the above table, objects are dropped onto the table, physics is stepped forward for several seconds so object settle, and then 20 random camera views are taken. Each object present in each view is saved as a prediction instance, with RGBD, mask, and model information. Camera views with no objects visible are discarded.

[1] Selected shapenet training categories: bag, traveling bag, travelling bag, grip, suitcase, birdhouse, bottle, bowl, camera, photographic camera, can, tin, tin can, cap, clock, computer keyboard, keypad, dishwasher, dish washer, dishwashing machine, helmet, jar, knife, laptop, laptop computer, microwave, microwave oven, mug, pillow, remote control, remote, telephone, phone, telephone set, cellular telephone, cellular phone, cellphone, cell, mobile phone. Test categories: display, video display, loudspeaker, speaker, speaker unit, loudspeaker system, speaker system, washer, automatic washer, washing machine, printer, printing machine, ashcan, trash can, garbage can, wastebin, ash bin, ash-bin, ashbin, dustbin, trash barrel, trash bin.

\section{Cluttered Manipulation Benchmark}

We generate 2574 manipulation tasks: 26 objects $\times$ three tasks (grasping, pushing, and rearrangement) $\times$ eleven levels of target object visibility $\big([0.0-0.1), [0.1-0.2),...,[0.9-1.0), 1.0) \big)$ $\times$ 3. We generated each task by first placing the target manipulation object at a fixed location. We then added distract objects directly in front of the target object, drawing distractor object y positions from a normal distribution with a center at the target object position and a y standard deviation of 0.1m, until a distractor object was added that caused the desired level of occlusion. We then added more distractor objects using the randomization parameters in the table below.

\textbf{grasping}: In this task the robot needed to grasp and object and lift it at least 5cm off the table.

\textbf{pushing}: In this task the robot needed to push an object 30cm to the right and 5cm forward. The robot succeed if the object ended within 2.5cm of the target position.

\textbf{rearrangement}: In this task the robot needed to grab and object and move it 20cm forward and 20cm to the left. The robot succeed if the object ended within 2.5cm of the target position.

\begin{center}
\begin{tabular}{| c | c |} \hline
 distractor object type & sampled from selected unseen objects [2]  \\ \hline
 number objects & uniform(1,7) \\ \hline
 object color & uniform random \\ \hline
 distractor object scale & uniform(0.5, 2) \\ \hline
 object rotation & upright, uniform random rotation \\ \hline
 distractor object position std dev & 0.25m \\ \hline
 \end{tabular}
\end{center}

[2] Target manipulation objects:\\
YCB Objects: master chef can, cracker box, sugar box, tomato soup can, mustard bottle, apple, orange, pitcher base, bleach cleanser, bowl, mug, wood block, tennis ball, rubiks cube \\
Objects from internet repository: cup (x2), glass (x3), vase, lamp (x5), trophy 

\section{MPPI Parameters}
We used the following parameters for MPPI:

\begin{center}
\begin{tabular}{| c | c |}\hline
 steps & 200 \\ \hline
 paths explored per step & 25 \\ \hline
 path length & 5 \\ \hline
 timestep length & 0.02 \\ \hline
 \end{tabular}
\end{center}

We used a stateful reward function for MPPI. The reward function had three states: \texttt{ungrasped}, \texttt{grasping}, and \texttt{grasped}. The state starts as \texttt{ungrasped}, and changes to \texttt{grasping} when \\
\\
\texttt{ungrasped=number steps<10 or (distance(hand, target object)>2.5mm and distance(hand, target object) decreased in the last 10 steps)}\\
\\
becomes false. When the state becomes \texttt{grasping}, the robot will close its gripper for 75 timesteps, and then the state will become \texttt{grasped}.

\texttt{ungrasped} and \texttt{grasping}: \texttt{Reward=-distance(robot hand, target object)-5*(hand height-38cm)-0.5*angle(robot hand, z axis)}\\

\texttt{grasped: Reward=-distance(robot hand, target object)-distance(target manipulation object position, current manipulation object position)}

\end{appendices}
\end{document}